\pdfoutput=1

\documentclass[11pt]{article}

\PassOptionsToPackage{dvipsnames,xcdraw}{xcolor}

\usepackage{xurl}
\usepackage[hyperref]{ACL}
\usepackage{times}
\usepackage{latexsym}

\usepackage[T1]{fontenc}

\usepackage[utf8]{inputenc}

\usepackage{xfrac}
\usepackage{anyfontsize}

\usepackage[
    noapacite,
    nohyperref,
    hyperrefmax,
    notextfont,
    nomathfont,
    nolineno,
    todo
]{mnn_style}

\usepackage{amssymb}
\usepackage{siunitx}

\newcommand*{\discard}[1]{}

\newcommand*{\WordI}{\symb{W}}
\newcommand*{\CharI}{\symb{L}}
\newcommand*{\Word}{w}
\newcommand*{\Char}{\ell}

\newcommand*{\BDER}{\fun{BDER}}
\newcommand*{\PDER}{\fun{PDER}}
\newcommand*{\RDER}{\fun{REDER}}
\newcommand*{\mtEff}{SU}
\newcommand*{\mtRDER}{RE-DER}
\newcommand*{\mEff}{\fun{\mtEff}}
\newcommand*{\ftDER}[1]{\textnormal{\textsuperscript{\textdagger}}#1}
\colorlet{sent}{red}
\colorlet{word}{blue}
\newcommand*{\WordSuffix}{\text{word}}
\newcommand*{\SentSuffix}{\text{sent}}
\newcommand*{\predictiveFunc}{f}
\newcommand*{\modelOutput}{y}
\NewDocumentCommand{\fWord}{oO{word}}{%
\IfNoValueTF{#1}
    { {\textcolor{#2}{\predictiveFunc_\WordSuffix}} }
    {\eqnmark[#2]{#1}{\predictiveFunc_\WordSuffix}}%
}
\NewDocumentCommand{\fSent}{oO{sent}} {%
\IfNoValueTF{#1}
    { {\textcolor{#2}{\predictiveFunc_\SentSuffix}} }
    {\eqnmark[#2]{#1}{\predictiveFunc_\SentSuffix}}%
}
\newcommand*\Wordi[1][wordi]{\!\eqnmark[red]{#1}{\Word_i}\!}
\newcommand*\Charj[1][charj]{\!\eqnmark[blue]{#1}{\Char_j \in \Word_i}\!}
\newcommand*\WordC[1][wordi]{\br{\Wordi[#1]}}
\newcommand*\CharC[1][charj]{\br{\Charj[#1]}}
\newcommand*{\pWord}{\mathcolor{word}{\Tilde{\modelOutput}^{\WordSuffix}}}
\newcommand*{\pSent}{\mathcolor{sent}{\Tilde{\modelOutput}^{\SentSuffix}}}
\newcommand*{\PWord}{\mathcolor{word}{{\modelOutput}^{\WordSuffix}}}
\newcommand*{\PSent}{\mathcolor{sent}{{\modelOutput}^{\SentSuffix}}}
\newcommand*{\fDiacPartial}{\fun{CCPD}}

\newcommand*{\selectedCorpus}{\train_\texttt{S}}
\newcommand*{\leftCorpus}{\train_\texttt{U}}
\newcommand*\modelD{\modelname{D2}}
\newcommand*\modelTD{\modelname{TD2}}

\let\ogendabstract\endabstract
\usepackage{arabtex}
\usepackage{utf8}
\newcommand\aratext[2][]{{\setcode{utf8}\RL{#2}}}
\let\endabstract\ogendabstract

\title{
A Context-Contrastive Inference\\
Approach To Partial Diacritization

}

\author{Muhammad ElNokrashy \\
  Microsoft \\
  Egypt \\
  \texttt{muelnokr@microsoft.com} \\
  \And
  Badr AlKhamissi \\
  EPFL \\
  Switzerland \\
  \texttt{badr.alkhamissi@epfl.ch}
}

\begin{document}

\maketitle

\begin{abstract}
Diacritization plays a pivotal role in improving readability and disambiguating the meaning of Arabic texts.
Efforts have so far focused on marking every eligible character (Full Diacritization).
Comparatively overlooked, Partial Diacritzation (PD) is the selection of a subset of characters to be annotated to aid comprehension where needed.
Research has indicated that excessive diacritic marks can hinder skilled readers---reducing reading speed and accuracy.
We conduct a behavioral experiment and show that partially marked text is often easier to read than fully marked text, and sometimes easier than plain text.
In this light, we introduce Context-Contrastive Partial Diacritization ($\fDiacPartial$)---a novel approach to PD which integrates seamlessly with existing Arabic diacritization systems.
$\fDiacPartial$ processes each word twice, once with context and once without, and diacritizes only the characters with disparities between the two inferences.
Further, we introduce novel indicators for measuring partial diacritization quality to help establish this as a machine learning task.
Lastly, we introduce \modelname{TD2}, a Transformer-variant of an established model which offers a markedly different performance profile on our proposed indicators compared to all other known systems.%
\footnote{Demo: \url{https://huggingface.co/spaces/bkhmsi/Partial-Arabic-Diacritization}}

\end{abstract}

\section{Introduction}

\begin{figure}
    \centering
    \includegraphics[width=1\linewidth]{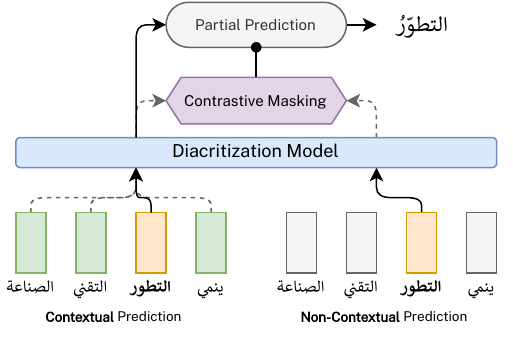}
\caption{
The proposed system employs a contextual diacritization model in two modes.
\figleft{} Model receives a word with its surrounding context,
\figright{} Model receives the word in isolation.
\textit{(Top)} The outputs are contrasted to select a subset of the text to diacritize.
}
\label{fig:main_figure}
\end{figure}

The Arabic language is central to the linguistic landscape of over 422 million speakers. It plays a pivotal role in the religious life of over a billion Muslims \cite{mijlad2022comparative}. As in other impure \newterm{abjad} writing systems, the Arabic script omits from writing some phonological features, like short vowels and consonant lengthening. This can affect reading efficiency and comprehension. Readers use context from neighbouring words, the domain topic, and experience with the language structure to guess the correct pronunciation and disambiguate the meaning of the text.

The Arabic NLP community has noticeably focused on the task of \oldterm{Full Diacritization} (\newterm{FD})---the modeling of diacritic marks on every eligible character in a text
(for example: \citealp{darwish17, mubarak19-highly, alkhamissi-etal-2020-deep}).
This is especially useful in domains where ambiguities are not allowed, or where deducing the correct forms might pose challenges for non-experts. Such domains may include religious texts, literary works like poetry, or educational material.

There are benefits for human readers, like facilitating learning. However, prior research suggests that extensive diacritization can inadvertently impede skilled reading by increasing the required processing time \cite{Haitham16, Ibrahim2013ReadingIA, aula14, Midhwah2020ArabicDA, ROMAN1987431, Hermena15}.
Nevertheless, diacritics are important morphological markers, even when excessive, and may benefit automated systems in language modeling, machine translation (MT), part-of-speech tagging, morphological analysis, acoustic modeling for speech recognition, and text-to-speech synthesis. For an example in MT, see \citet{fadel-etal-2019-neural,habash-etal-2016-exploiting,alqahtani-etal-2016-investigating,diab-etal-2007-arabic}.

\begin{table}[!t]
\centering
\begin{tabular}{@{}c@{\hskip 0.2em}cr@{}}
\toprule
&
    \textbf{System} & \textbf{Sentence} \\
\midrule
& \multicolumn{2}{c}{Full Diacritization} \\
\multicolumn{2}{c}{\textbf{Truth}}
                     & {\aratext{سَوْفَ نَحْيَا هُنَا .. سَوْفَ يَحْلُو النَّغَمُ}} \\
\midrule
\multirow{4}{*}{\rotatebox[origin=c]{90}{--MV, hard--}}
& \multicolumn{2}{c}{Partial Diacritization} \\
    & \modelname{TD2} & {\aratext{سوفَ نحيا هُنا .. سوفَ يحلو النغَمْ}} \\
    & \modelname{D2}  & {\aratext{سوف نحيَا هنا .. سوفَ يحلو النُّغْمَ}} \\
\midrule
\multirow{4}{*}{\rotatebox[origin=c]{90}{---\modelname{D2-SP}---}}
    & hard            & {\aratext{سوف نحيا هنا .. سوفَ يحلو النغم}} \\
    & soft $>0.1$~    & {\aratext{سوف نحيَا هنا .. سَوف يحلو النَّغم}} \\
    & soft $>0.01$    & {\aratext{سَوف نحيَا هنا .. سَوْفَ يحلو النَّغم}} \\
\bottomrule
\end{tabular}
\caption{
System Outputs on text from the \dataset{Tashkeela} testset.
``$\fun{soft}$'' methods have an adjustable threshold. \textbf{MV} uses majority-voting, while \textbf{SP} (Single-Pass) doesn't (see \cref{sec:majority-voting,sec:onepass}).
The line translates into: ``We will live here. The singing will be sweet.''
}
\label{tab:examples}
\end{table}

\subsection{Contributions}
\paragraph{Context-Contrastive Partial Diacritization}
We propose a novel method for \oldterm{Partial Diacritization} (\newterm{PD}) which seamlessly utilizes existing Arabic FD systems.
We exploit a statistical property of Arabic words wherein readers can guess the correct reading of most unmarked words with \emph{minimal context}.
To select the letters to mark, $\fDiacPartial$ sees each word twice: (1) within its sentence context, and (2) as an isolated input with no context. The two predictions are combined to retain only those diacritics which present comparatively new information that may aid reading comprehension (\cref{sec:partial-diac}).

\paragraph{Human Evaluation}
We conduct a behavioral experiment to compare ease of reading for text with all, some, or no diacritics. Diacritics are selected via a neural model mask to simulate native partial marking. The results support prior work on the effect of different degrees of diacritcs on reading efficiency and comphrension, and is the motivation for this work (see \cref{sec:human-eval}).

\paragraph{Performance Indicators}
We introduce a set of automatic indicators to gauge the performance and usefulness of our method on several public models, in light of the scarcity of supervised labeled partially diacritized datasets (see \Secref{sec:new-metrics}).

\paragraph{Transformer D2}
And last, we present the \textbf{\modelname{TD2}} model, a Transformer variant of \modelname{D2} \cite{alkhamissi-etal-2020-deep}, which shows markedly improved Partial~DER performance at \SI{5.5}{\%} compared to \SI{11.2}{\%}, even while marking a \emph{larger} percentage of text at \SI{24.6}{\%} compared to \SI{6.5}{\%} for \modelname{D2} (see \Secref{sec:td2}).


\begin{table}[t]
\begin{center}
\begin{tabular}{@{}cllll@{}}
\toprule
\bf Glyph & \bf Name & \bf Type & \bf BW & \bf IPA \\
\midrule 
{\aratext{هُ}}  & dammah      & \d{h}arakāt  & \it u  & /hu/ \\
{\aratext{هَ}}  & fathah      & \d{h}arakāt  & \it a  & /ha/ \\
{\aratext{هِ}}  & kasrah      & \d{h}arakāt  & \it i  & /hi/ \\
{\aratext{هْ}}  & sukūn       & sukūn        & \it o  & /h./ \\
{\aratext{هٌ}}  & dammatain   & tanwīn       & \it N  & /hun/ \\
{\aratext{هً}}  & fathatain   & tanwīn       & \it F  & /han/ \\
{\aratext{هٍ}}  & kasratain   & tanwīn       & \it K  & /hin/ \\
{\aratext{هّ}}  & shaddah     & shaddah      & $\sim$ & /h:/ \\
\bottomrule
\end{tabular}
\end{center}
\caption{Primary Arabic Diacritics on the Arabic letter \textit{h}. \textbf{BW} is the Buckwalter transliteration of the vowel or syllable. Adapted from \citet{alkhamissi-etal-2020-deep}.}
\label{tab:diac-table}
\end{table}

\section{Motivation}
\label{sec:motivation}

Previous research underscores the substantial influence of diacritization on the reading process of Arabic text. Skilled readers exhibit a tendency to read highly diacritized text at a slower pace compared to undiacritized. The effect on reading speed and accuracy has been substantiated across numerous studies involving diverse age groups and linguistic backgrounds \cite{Haitham16, Ibrahim2013ReadingIA, aula14, Midhwah2020ArabicDA, ROMAN1987431, Hermena15, VariationDiacritics22}.
In addition, studies monitoring eye movements during reading link the increased reading times for diacritized text to an increase in fixation frequency and duration---a key indicator of the word identification process \cite{ROMAN1987431, Hermena15}.
\emph{Partial Diacritization} may thus lead to a better reading experience for a wide range of readers, including in cases with dyslexia or visual impairments.

\pagebreak

\section{Arabic Preliminaries}

\paragraph{Diacritics in Modern Arabic Orthography}
The inventory of diacritics used in modern typeset Arabic comprises at least four functional groups. First are \d{h}arakāt and sukūn, which indicate vowel phonemes or their absence. Then tanwīn and shaddah indicate case-inflection morphemes and consonant lengthening.
There are other marks, e.g. for recitation features in the Qur'an. Only the marks in \BTableref{tab:diac-table}, part of modern Arabic orthography, are discussed here.
\d{H}arakāt are vowel diacritics that provide vowel information. Sukūn indicates the absence of a vowel, which indicates consonant clusters or diphtongs.
Tanwīn diacritics indicate the phonemic pair of a vowel and the consonant /n/ at the end of words, serving case inflection purposes.
These diacritics play essential roles in Arabic orthography and pronunciation \cite{VariationDiacritics22}.

\begin{figure}
\centering
    \includegraphics[width=1\linewidth]{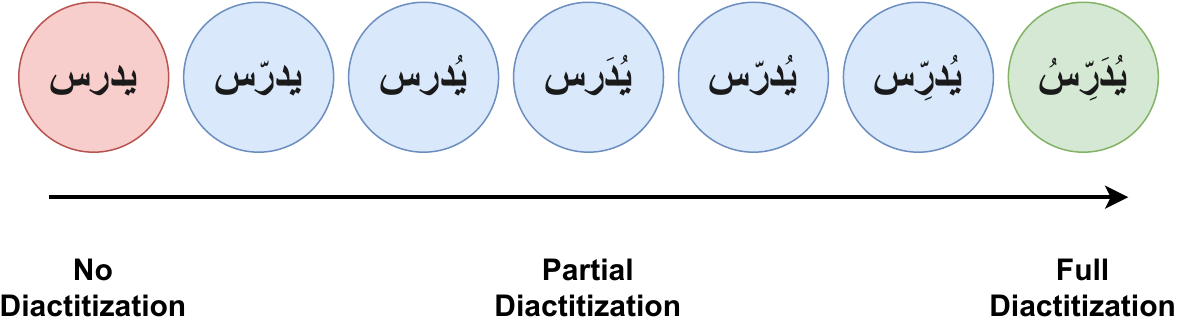}
\caption{Seven possible ways among others of writing the word \code{yudarrisu} (English: He teaches) with varying levels of diacritc coverage.}
\label{fig:examples}
\end{figure}

\paragraph{Modes of Diacritization.}
Since diacritics are optional and their usage can vary widely---practical patterns have emerged.
\citet{VariationDiacritics22} identifies seven modes of diacritization. These modes can be ordered based on the quantity of diacritics used, ranging from no diacritization to complete diacritization.
Deeper levels, like \textit{complete} diacritization, are less frequently used than shallower levels (no diacritization, and so on).
Examples in \cref{fig:examples}.


\section{Methods}
\label{sec:partial-diac}

\subsection{Full Diacritization Models}
We utilize deep neural sequence models for character diacritization.
We test the pretrained models \modelname{D2}, \modelname{Shakkelha} and \modelname{Shakkala} \cite{alkhamissi-etal-2020-deep,fadel19,shakkala}.
Further, we introduce the \modelname{TD2} model: a Transformer variant of the LSTM-based \modelname{D2}.
A description of \modelname{TD2} and the \modelname{D2} architecture can be found at \cref{sec:td2}.

\pagebreak
\subsection{What Context Do Models See?}

\paragraph{Context extraction.}
Let $\fun{ctxt}(x; \mathbb{X})$ contextualize $x$ from $\mathbb{X}$.
Then the word and letter context extractors are functions which use the surrounding words and letters around positions $i$ and $j$.

Let the sub-sequence $s_{i,T}$ contain all the words in a \emph{segment} of length $2T$ s.t. $s_{i,T} =\{w_t \mid t \in i \pm T\}$ as used in training \modelname{D2}/\modelname{TD2}.

\begin{alignat}{5}
\br[\big]{\Wordi[Wordi]{}}_T &=
    \func{ctxt}(\Word_i&&;
        \{ \Word \ldots &&\in s_{i,T} &&\}
    )
\\
\br[\big]{\Charj[Charj]{}}\hphantom{{}_T} &=
    \func{ctxt}\pa{ \Char_j&&;
        \{ \Char \ldots &&\in \Wordi &&\}
    }
\end{alignat}
\annotate[yshift= 0.4em]{above,right}{Wordi}{Words around position $i$}
\annotate[yshift=-0.2em]{below,right}{Charj}{Letters in Word $i$}
\vspace{0em}

\noindent
Thus $\WordC_T$ refers to a segment containing the word $i$, while $\CharC$ refers to the intra-word context for letter $j$ in word $i$.

\paragraph{Contextual Prediction} sees the full segment.
\begin{equation}
\label{eq:contextual-application}
    \fSent(i, j) = f\big(
        \discard{\WordI}{\WordC_T},
        \discard{\CharI}{\CharC}
    \big)
\end{equation}


\paragraph{Single Word Prediction} sees only the current word\footnotemark{} at position $i$. Note the $w_i$ instead of $\br{w_i}_T$.
\footnotetext{Sentences are split on spaces. Words are not tokenized.}
\begin{equation}
\label{eq:non-contextual-application}
    \fWord(i, j) = f\big(
        \discard{\WordI}{\hphantom{\big\{} \Wordi \hphantom{\big\}_T}},
        \discard{\CharI}{\CharC}
    \big)
\end{equation}
Thus the two modes are defined as follows: $\fWord$ is single-word application, while $\fSent$ uses sentence context. Both use the same model $f$.



\subsection{\textit{C}ontext-\textit{C}ontrastive \textit{P}artial \textit{D}iacritization}
\label{sec:ccpd}

The function $\fWord$ takes each word in isolation. If its prediction is correct, we assume that this reading is common and easy to guess by a human reader.
By contrast, $\fSent$ utilizes word context \emph{within a sentence}. If a word has multiple possible readings which the sentence disambiguates, we expect $\fSent$ with sentence context to out-perform $\fWord$.

Following from these premises, we propose the following method to combine both sources of information to diacritize text partially, with justification.

\paragraph{CCPD: Algorithm}
Using $\fSent$ and $\fWord$ predictions for a word and letter $(\Word_i, \Char_j)$, the system assigns or omits a diacritic according to $\fDiacPartial(i,j)$:
\begin{equation}
\begin{aligned}
\label{eq:pdd-assign}
        &\fDiacPartial(i, j) = \\
        &\quad\;
            \begin{cases}
                \PSent(i, j)  & \gets \func{mark}(i,j), \\
                \varnothing   & \gets \text{otherwise}
            \end{cases}
\end{aligned}
\end{equation}


In \textbf{hard} or \textbf{disagreement} mode, the letters where inferences with and without context \emph{agree} are left bare (regardless of the correctness of the prediction; unknown during inference). Otherwise, the contextual prediction $\PSent$ is returned.

\vspace{-1.5em}
\begin{flalign}
\label{eq:pdd-mark-hard}
    \func{mark}_\text{hard}(i, j) &:=
        \PSent_{i,j} \neq \PWord_{i,j}
\end{flalign}

In \textbf{soft} or \textbf{confidence} mode, only the letters where the logit $\pSent$ exceeds the logit $\pWord$ by a margin $\theta$ receive a diacritical mark.

\vspace{-1.5em}
\begin{flalign}
\label{eq:pdd-mark-soft}
    \func{mark}_\text{soft}(i, j) &:=
        \pSent_{i,j} > \theta + \pWord_{i,j}
\end{flalign}

\section{Performance Indicators}
\label{sec:new-metrics}
To measure performance on the partial diacritization task, we propose four indicators as approximate gauges for model alignment with human expectation. These indicators aim to address the challenge of the limited availability of high-quality partially diacritized test datasets, providing valuable insights into the model's capabilities.

Indicators may use the marked ($\selectedCorpus$) and unmarked ($\leftCorpus$) subsets of the corpus ($\train$).
\begin{equation}
\begin{aligned}
    \selectedCorpus &= \big\{
        i,j \in \train \mid \hphantom{\lnot} \func{mark}(i, j)
    \big\} & \textcolor{Gray}{\text{selected}}
    \\
    \leftCorpus &= \big\{
        i,j \in \train \mid \lnot \func{mark}(i, j)
    \big\} & \textcolor{Gray}{\text{unmarked}}
\end{aligned}
\end{equation}

\subsection{SR: Selection Rate}
\label{sec:selection}
\begin{equation}
\label{eq:sr}
    \eb[Red]{}{\!\fun{SR}}(\train) =
        \frac{
            \abs{ \discard{\eb[Gray]{selected}}{\selectedCorpus{}} }
        }{
            \abs{ \discard{\eb[Gray]{corpus}}{\train} }
        }
\end{equation}
\textbf{SR} is the proportion of characters assigned a diacritic by \Funcref{eq:pdd-assign}.
Literature shows partial diacritization hovers around \SI{1.2}{\%}--\SI{9.5}{\%} coverage in some professionally published books \citep{VariationDiacritics22}, while other research suggests that deliberate partial diacritization by native speakers results in a rate around \SI{19}{\%}--\SI{26}{\%} \citep{Esmail22}.

\subsection{Scoped Diacritic Error Rates DER(f, D)}
\label{sec:scoped-der}
\vspace{1.1em}
\begin{equation}
\label{eq:sder}
\func{DER}(
        \eb[NavyBlue]{system1}{f},
        \eb[Emerald]{corpus1}{\train}
    ) =
    \frac{1}{ |\eb[Emerald]{corpus2}{\train}| } 
    \sum_{ \eb[Gray]{index1}{i,j\in\train} }
    \Iv{
        \eb[NavyBlue]{system2}{f}
            \neq
        \eb[OliveGreen]{ground-truth}{\func{gt}}
    }_{\eb[Gray]{index2}{i,j}}
    \;\;
\vspace{0.5em}
\end{equation}
DER in traditional literature is the ratio of diacritics erroneously predicted over all eligible letters in a corpus. Trivially, we parameterize DER by the system $f$ and corpus (subset) $\train$.
\annotatetwo[yshift= 0.5em]{above,left}{system1}{system2}{Predictor}
\annotate   [yshift= 0.8em]{above,left}{ground-truth}{True labeler}
\annotate   [yshift=-0.4em]{below,left}{corpus2}{(Subset of) letters}
\annotate   [yshift=-0.2em]{below,right,label below}{index1}{Word $i$, char $j$}

\subsubsection{P-DER: Partial Diacritic Error Rate}
\label{sec:partial-der}
\vspace{1.0em}
\begin{equation}
\label{eq:pder}
\PDER(\train) =
    \func{DER}(
        \fSent[context]{},
        \eb[Gray]{corpus}{\selectedCorpus{}}
    )
\;\;\;\;
\end{equation}
We calculate the DER on $\selectedCorpus$, which includes all characters in the corpus assigned diacritics by $\fun{CCPD}$, following \Funcref{eq:pdd-assign}.
\annotate[yshift= 0.2em]{above,left}{context}{Contextual prediction}
\annotate[yshift= 0.2em]{above,right}{corpus}{Marked letters}

\subsubsection{B-DER: Basic Diacritic Error Rate}
\label{sec:basic-der}
By the intuition in \cref{sec:ccpd} regarding the non-contextual predictions of $\fWord$,
we calculate the DER on the whole corpus $\train$
and use \newterm{B-DER} as a proxy for human error on plain unmarked text.

\vspace{0.2em}
\begin{equation}
\label{eq:bder}
    \eb[Green]{}{\!\BDER}(\train) = \func{DER}(\fWord[noncontext]{}, \eb[Gray]{corpus}{\train})
\annotate[yshift= 0.2em]{above,left}{noncontext}{Non-Contextual prediction}
\annotate[yshift= 0.2em]{above,right}{corpus}{All letters}
\end{equation}


\subsection{RE-DER: Reader DER}
\label{sec:general-error}
We combine the Partial DER and Basic DER into one general measure of the total error in a partially diacritized text.
The Basic DER indicator is used as a proxy which may be taken as an upper bound for mistakes made by an experienced reader.
Formally, this indicator incorporates the error of the non-contextual model for the unmarked text, and of the contextual-model for the marked text (because the system explicitly annotates it).
Thus \newterm{RE-DER} (Reader DER) is:
\begin{equation}
\label{eq:r-der}
\begin{aligned}
    \eb[Plum]{}{\!\RDER}(\train)
        &= (\func{SR})     \hspace{-1em} && \times \fun{DER}({\fSent}, \selectedCorpus)
    \\
        &+ (1 - \func{SR}) \hspace{-1em} && \times \fun{DER}({\fWord}, \leftCorpus)
\end{aligned}
\end{equation}

\subsection{\mtEff{}: Signal Utilization}
\label{sec:general-improvement}
How well does $\fDiacPartial$ utilize its marked subset to disambiguate the text?
We gauge how informative the added diacritics are by measuring the change in DER within the marked subset of text.
\mtEff spans $[-1, 1]$  where positive values denote improvement over no annotation.
\newterm{\mtEff} is defined as:
\begin{flalign}
\label{eq:signal-utilization}
\begin{split}
    \mEff(\train)
    &= \frac{
            \eb[Green]{}{\!\BDER} - \eb[Plum]{}{\!\RDER}
        }{
            \eb[Red]{}{\!\fun{SR}}
        }
\end{split}
\end{flalign}
\textit{E.g.} an \mtEff{}=50\% shows that half of the produced annotations clarify ambiguities (lower \mtRDER{}).

\def\p{}
\begin{table*}[!ht]
\centering
\begin{tabular}{@{}ll|r|r|rrr|rr@{}}
\toprule
    & \bf System
    & \multicolumn{1}{c|}{\bf SR}
    & \multicolumn{1}{c|}{\bf \mtEff}
    & \multicolumn{1}{c} {\bf P-DER}
    & \multicolumn{1}{c} {\bf \mtRDER{}}
    & \multicolumn{1}{c} {\bf B-DER}
    & \multicolumn{1}{|c}{\bf DER  }
    & \multicolumn{1}{c} {\bf WER  }
\\
\midrule
    & \multicolumn{1}{l|}{\textsc{Reference Range}}
        & $13\pm 8\ \%$
    & \multicolumn{1}{c|}{$\uparrow \pm 100\%$}
    & \multicolumn{5}{c} {$\downarrow 100\% \ldots 0\%$}
\\
\midrule
        & \citealp{shakkala}
                &      8.6\p &     70.9 
                & 17.9\p &     \ftDER{3.6\p} &   9.7\p &     3.58\p &    11.19\p \\
        & \citealp{fadel-etal-2019-neural} (big)*
                &      8.9\p & \bf 86.5 
                & 7.8\p & \ftDER{\bf 1.6\p} &   9.3\p & \bf 1.60\p & \bf 5.08\p \\
        & \citealp{alkhamissi-etal-2020-deep} - \bf\modelD{}
                &      6.5\p & \it 81.5 
                & 11.2\p & \ftDER{\it 1.8\p} &   7.1\p & \it 1.85\p & \it 5.53\p \\
\cmidrule[0.01pt]{1-9}
    \multirow{7}{*}{\rotatebox[origin=c]{90}{---Ours---}}
        & \textbf{\modelD{}} (SP, hard)
                &      6.8\p & \bf 75.0 
                & 15.0\p &  \ftDER{\bf  2.0\p} &
                \multirow{5}{*}{\rotatebox[origin=c]{45}{7.1\p}}  &
                \multirow{5}{*}{\rotatebox[origin=c]{45}{2.00\p}} &     \multirow{5}{*}{\rotatebox[origin=c]{45}{6.42\p}} \\
        & \textbf{\modelD{}} (SP, soft $> 0.4$)
                &      5.9\p &     47.5 
                & \bf  3.1\p &            4.3\p &          &            &            \\
        & \textbf{\modelD{}} (SP, soft $> 0.2$)
                & \bf 11.5\p &     37.4 
                & \it  4.7\p &            2.8\p &          &            &            \\
        & \textbf{\modelD{}} (SP, soft $> 0.1$)
                & \bf 15.2\p &     31.6 
                & 5.1\p &            2.3\p &          &            &            \\
        & \textbf{\modelD{}} (SP, soft $> 0.01$)
                &     21.8\p &     23.4 
                & 4.9\p &  \bf       2.0\p &          &            &            \\
\cmidrule[0.01pt]{2-9}
        & \textbf{\modelTD{}} (MV, hard)
                & \it 24.6\p & \bf 91.5 
                & \it  5.5\p & \ftDER{\it  2.4\p} &   25.0\p &     2.44\p &     7.68\p \\
\bottomrule
\end{tabular}
\caption{
Partial Diacritics Results on \dataset{Tashkeela} via the indicators outlined in \cref{sec:new-metrics}. \textbf{SR} is Selection Rate, \textbf{P-DER} is Partial Diacritic Error Rate, \textbf{\mtRDER} is Reader DER, while \textbf{\mtEff} is Signal Utilization (eq. \ref{eq:signal-utilization}).
We report DER and WER as well.
\textbf{(SP)} (SinglePass) results run inference on the whole sentence at once, with no segmentation or majority voting.
\quad\citet{fadel19}* uses the \modelname{lstm-big-20} configuration and uses extra training data beyond the \dataset{Tashkeela} \texttt{train} split; it is part of the public code release, but is not reported on in the original paper.
\quad\textdagger{} Under $\fun{hard}$ mode letter selection, \mtRDER{} is equal to $\fun{DER}(\fSent, \train)$, which follows from eq. (\ref{eq:pdd-mark-hard}) \& (\ref{eq:r-der}).
\quad
\emph{A lower B-DER is preferable (indicating a stronger base). While lower P-DER makes for smoother reading (fewer annotation errors). A natural SR is around 13\%.}
}
\label{tab:test_results}
\end{table*}

\section{Experimental Modeling}
\subsection{TD2: Transformer D2}
\label{sec:td2}
The \modelname{TD2} model is a Transformer adaptation of \modelname{D2} \citep{alkhamissi-etal-2020-deep}.
It comprises two encoders of 2 layers each, for tokens and characters.
Both models use the same architecture.
The feature and intermediate widths are (768, 2304).
The token model is initialized from the pretrained weights of layers 1-2 of \textit{CAMeL-Lab/bert-base-arabic-camelbert-mix-ner} \citep{camel-lm} as provided by HuggingFace\footnote{HuggingFace: \citet{wolf2020huggingfaces}}.
The character model is initialized from layers 3-4 of the same pretrained weights.

\subsection{SP: Single-Pass Inference}
\label{sec:onepass}
For recurrent (non-Transformer) models like \modelname{D2}, we also test inference without segmentation and majority voting. Each sentence is passed in full the model, once. This can save inference time (by avoiding overlapping windows), but sacrifices some reliability (as the model had been trained on smaller window sizes).

\subsection{Other Models}
\modelname{Shakkala} \citep{shakkala} and \modelname{Shakkelha} \citep{fadel-etal-2019-neural} are LSTM-based models that view sentences as a sequence of characters. In contrast, \modelname{D2} and \modelname{TD2} view sentences as word sequences.


\section{Results}

\subsection{Full Diacritization Performance}
Following prior work, we report the Diacritic Error Rate (DER) and Word Error Rate (WER) including case-endings which are located at the word's end, usually determined by the word's syntactic role. Predicting these diacritics is more challenging compared to core-word diacritics, which specify lexical properties and lie elsewhere within the word.

\subsection{Partial Diacritization Performance}

\paragraph{Intuition}
\label{sec:indicator-intuition}
Overall it is desirable to observe: (1) a natural SR value close to native human annotation rates; (2) a low B-DER close to the SR, signifying a capable model which allows a clean partial annotation of only the hardest letters, (3) a low P-DER to signify high accuracy in the committed diacritics, and (4) a low \mtRDER{} which gauges the overall expected reading experience given the partial annotation and the expected guessing error. \textit{Overall, this corresponds to a balance between a \textbf{high SU}, a \textbf{natural SR}, and a \textbf{low P-DER}, roughly ordered.}

\paragraph{Model results}
Notice that the recurrent \modelname{D2} is conservative in its selection with \textbf{SR} at 6.46\% versus \modelTD{}'s 24.61\%.
\modelTD{} under-performs in non-contextual mode with \textbf{B-DER} at 25\% compared to 7.1\% by \modelD{}. Counter-intuitively, this leads it to a higher \textbf{SU} at 91.5\% as its contextual mode corrects 22.5 of every 25 errors its non-contextual mode commits.
\\
Both systems agree that much of the text (at least 75\%) is reasonably guessable by the reader without annotation, by \textit{taking B-DER as an upper-bound on human guess error}.
We may claim that between 75\% and 93\% of characters require no diacritization to disambiguate (the extremes of the two systems).
Both \modelD{} and \modelTD{} correct most of their B-DER error via $\fDiacPartial$, implying that most letters which had errors in non-contextual mode were accurately selected by $\fDiacPartial$ \emph{and} correctly predicted by $\fSent$, leading to an \mtRDER{} < B-DER.



\paragraph{Soft marking}
\textbf{SR} can be tweaked in soft selection mode. Let's analyze the model with SR=15.2\% (\cref{tab:test_results}) which provides a low P-DER at 5.1\% (few diacritics added are erroneous) and a low \mtRDER{} at 2.3\% (around 4.8 out of 7.1\% expected errors are corrected via the added diacritics). Notice the low SU (utilization) however, as only 31.6\% of the annotated letters contribute to the drop in error.
\\
Assuming that the selected letters are the hardest to guess for a reader, this suggests that \modelname{D2-SP} models can disambiguate the text by 39.4\% to 71.8\% (relative improvement in \mtRDER{}/B-DER) by marking only 5.9\% to 21.8\% of it.
While \mtRDER{} improves, P-DER and SU worsen, suggesting that increased annotation may give only marginal improvement in overall readability if the final model output is not perfectly clean.





\section{Behavioral Experiment on Partial Diacritization}
\label{sec:human-eval}

We start with the hypothesis that reading partially diacritized text may be easier than reading fully diacritized text, while providing some benefit in disambiguation or ease-of-reading.

\begin{figure}[h]
\centering
    \includegraphics[width=1\linewidth]{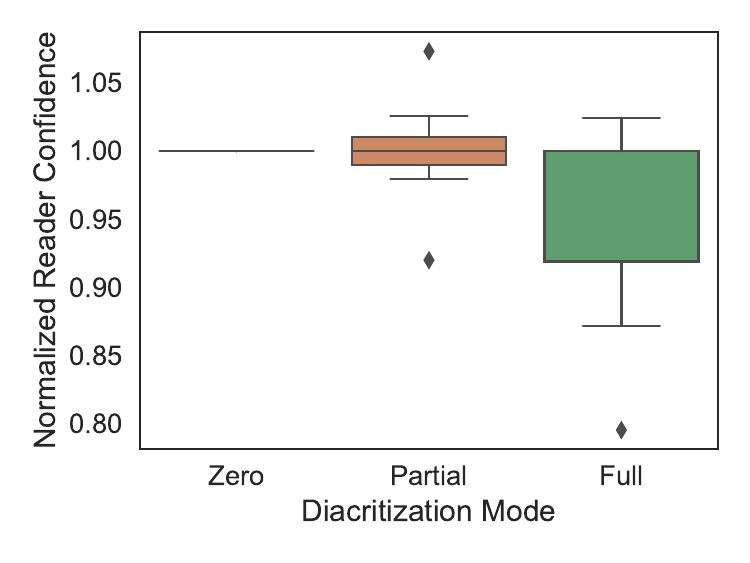}
\caption{
\textbf{Behavioral Experiment}: Self-reported scores for reading comprehension of sentences with Zero/Partial/Full diacritics.
Scores are aggregated per mode and normalized by the score of the Zero mode of each participant, making the results comparable across participants.
Notice the higher score points for partially diacritized samples (with some regressions), compared to the big regression for fully diacritized text.
}
\label{fig:human-eval}
\end{figure}

To test this hypothesis, we conduct a behavioral experiment using machine-predicted partial diacritization \emph{masks}.
The diacritical marks themselves are the true labels.
This is to ensure that the presented text is accurate, even if the model output is sub-optimal.
The question is then: \emph{Does $\fDiacPartial$ using $\fSent$ and $\fWord$ select useful letters to be diacritized?}
See the results in \BFigref{fig:human-eval}.

\subsection{Behavioral Experiment Setup}
\begin{figure*}
\centering
\begin{subfigure}[c]{0.32\linewidth}
        \includegraphics[width=1.0\linewidth]{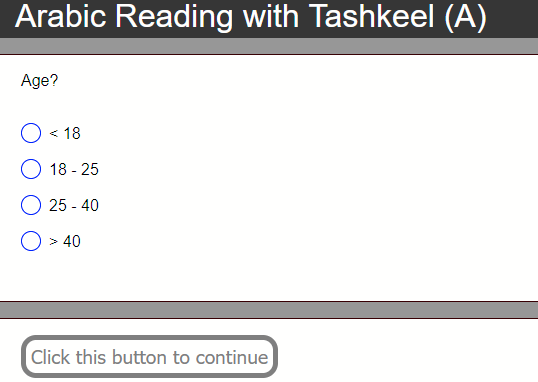}
    \caption{Demographic questions, like Age and Arabic Proficiency.}
    \label{fig:survey-demographic-questions}
\end{subfigure}\hfill%
\begin{subfigure}[c]{0.32\linewidth}
        \includegraphics[width=1.0\linewidth]{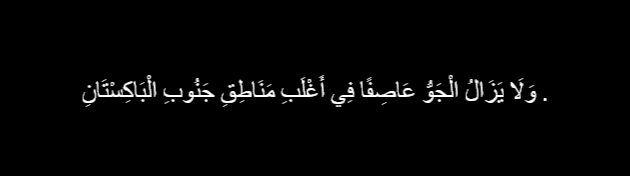}
    \caption{Examples are shown for 2-5 seconds with zero/some/all diacritics, in white typeface on black background.}
    \label{fig:survey-example}
\end{subfigure}\hfill%
\begin{subfigure}[c]{0.32\linewidth}
        \includegraphics[width=1.0\linewidth]{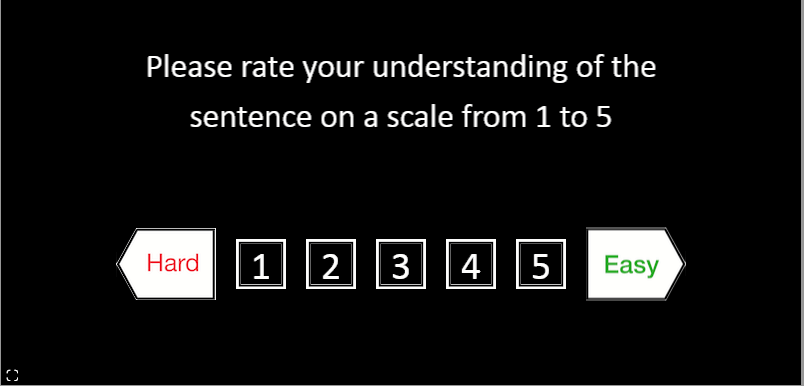}
    \caption{Participants have to socre their understanding of the previous sentene between 1 and 5.}
    \label{fig:survey-scoring}
\end{subfigure}
\caption{
Screenshots of Behavioral Experiment Steps.
(\ref{fig:survey-demographic-questions}) Demographic info is collected on: Age, Number of Languages Spoken, Arabic Proficiency, and Native Arabic Dialect Spoken.
(\ref{fig:survey-example}) Each participant was assigned one of 3 different test versions. All versions include the same sentences, but differ in which sentences are assigned how many diacritics.
(\ref{fig:survey-scoring}) Participants rate their understanding of an example from 1 to 5.
Participants were informed that the anonymized scores and metadata will be collected and used in an academic work.
}
\label{fig:survey-screens}
\end{figure*}

\paragraph{Demographics}
We utilize the PsyToolkit online platform\footnote{PsyToolkit: \citet{psytoolkit-2010,psytoolkit-2017}} to conduct a behavioral experiment aimed at assessing the impact of different levels of diacritization (modes) on reading speed and accuracy.
Data was gathered from a group of 15 participants, covering various age groups and native speakers of different Arabic dialects.
The majority of participants (10) fell within the 25--40 age group; 2 were below 25, and 3 were above 40.
All participants spoke at least 2 languages, while 6 spoke 3 or more.
Eight participants spoke either Gulf (4) or Maghrebi (4) Arabic natively, and 7 spoke Egyptian (2), Levantine (2), or Sudanese (2). One did not report a native dialect.
All participants reported being native speakers of Arabic.

\paragraph{Dataset}
To measure the impact of diacritics, we select 30 sentences from various domains in the \dataset{WikiNews} testset \citep{darwish17}.
Each sentence is presented in three variants with Zero/Partial/Full diacritics.
The Partial variant is produced using our $\fDiacPartial$ algorithm and the \modelname{D2-SP} model in $\fun{hard}$ selection mode to mask out easy or guessable ground truth diacritics.

\paragraph{Data Splits}
The data is split into 3 buckets of 10 sentences each.
We create 3 test sets such that the buckets are rotated and each appears exactly once in any mode (Zero/Partial/Full).
For example, the first bucket would appear in the test sets A, B, C in its Full, Zero, and Partial variants.
This is done to ensure that any participant sees a sentence exactly once, to avoid biasing their rating by repeated exposure. This ensures also that each sentence is seen equal times in each mode.

\paragraph{Timing and Scoring}
Participants are shown each sentence for a few seconds proportional to the word count of the sentence: $\sfrac{1}{4} ~ |\text{words}|$.
Then the participants are prompted to rate their comprehension  by a score from 1 to 5, where 1 indicates difficulty and 5 indicates ease in understanding the sentence.
The design intention is to measure the reading experience of a native speaker when scanning a non-technical text in Modern Standard Arabic.

\subsection{Findings}
\cref{fig:human-eval} illustrates the per-mode average self-reported reading comprehension scores, normalized by the score of the no diacritization (Zero) mode. The results indicate that Full diacritization hinders reading accuracy when participants are provided with a limited amount of time to read the text, aligning with findings discussed in \cref{sec:motivation}.
In contrast, the Partial diacritization mode sometimes enhances reading comprehension performance compared to the Zero mode.

The data from this behavioral experiment will be available on the paper's GituHub repository%
\footnote{GitHub: \href{https://github.com/munael/arabic-partial-diacritization-ccpd}{\texttt{munael/arabic-partial-diacritization}}}.





\section{Related Work}
The literature has dealth with both types of diacritization: Full (FD) and Partial (PD). The majority has focused on FD.
In PD, a distinct approach that seeks to augment reading comprehension by incorporating only the minimal requisite diacritics, we note a few works focus \citep{almanea2021automatic, mijlad2022comparative}.
Other works take a morphological analysis approach \citep{obeid-etal-2022-camelira,obeid-etal-2020-camel,alqahtani-etal-2016-investigating,shahrour-etal-2015-improving,habash-rambow-2007-arabic}.

\subsection{Full Diacritization}

\paragraph{Non-Neural Methods}
This class of methods combines linguistic rules with non-neural modelling techniques, such as Hidden Markov Models (HMMs) or Support Vector Machines (SVMs), which were widely employed over a decade ago. For example, \citet{elshafei06} utilize an HMM to predict diacritization based on bigram and trigram distributions. \citet{Bebah2014HybridAF} extract morphological information and utilize HMM modeling for vowelization, considering word frequency distribution. \cite{shaalan-etal-2009-hybrid} combine lexicon retrieval, bigram modeling, and SVM for POS tagging, addressing inflectional characteristics in Arabic text. \citet{darwish17} operate diacritization in two phases, inferring internal vowels using bigrams and handling case endings via an SVM ranking model and heuristics. \citet{Said2013AHA} follow a sequence-based approach, involving autocorrection, tokenization, morphological feature extraction, HMM-based POS tagging, and statistical modeling for handling OOV terms. \citet{zitouni09} approaches the problem using maximum entropy models.


\paragraph{Neural Methods}
Recent literature has focused more on neural-based systems. \citet{belinkov-glass-2015-arabic} were the first to show that recurrent neural models are suitable candidates to learn the task entirely from data without resorting to manually engineered features such as morphological analyzers and POS taggers.
\citet{alkhamissi-etal-2020-deep} used a hierarchical, BiLSTM-based model that operates on words and characters separately with a cross-level attention connecting the two---enabling SOTA task performance and faster training and inference compared to traditional models.
Many prior works have used recurrent-based models for the FD task \cite{Thubaity20,darwish2020arabic,Abandah20,fadel19,Boudchiche, Moumen2018EvaluationOG}, others used the Transformer architecture \cite{mubarak19-highly}, while others used ConvNet architectures \cite{alqahtani2019efficient}.

\paragraph{Hybrid Methods}
Some works have combined neural and rule-based or other methods to improve Arabic diacritization \citep{alqahtani2020multitask,hamza20,darwish2020arabic, Alqudah2017InvestigatingHA,Hifny18}.

\subsection{Partial Diacritization}
Similar to this work, previous research has explored the selective diacritization of Arabic text. \citet{Alnefaie17} harnessed the output of the MADAMIRA morphological analyzer \cite{pasha14-madamira} and leveraged WordNet to generate word candidates for diacritics. This work focused on resolving word ambiguity through statistical and contextual similarity approaches to enhance diacritization effectiveness. 
\citet{Alqahtani2019HomographDT} focused on selective homograph disambiguation, proposing methods to automatically identify and mark a subset of words for diacritic restoration. Evaluation of various strategies for ambiguous word selection revealed promising results in downstream applications, such as neural machine translation, part-of-speech tagging, and semantic textual similarity, demonstrating that partial diacritization effectively strikes a balance between homograph disambiguation and mitigating sparsity effects.
\citet{Esmail22} employ two neural networks to predict partial diacritics---one considering the entire sentence and the other considering the text read so far. Partial diacritization decisions are made based on disagreements between the two networks, favoring the prediction conditioned on the whole sentence.

\section{Discussion}

In this work, we have presented a novel approach for partial diacritization using context-contrastive prediction. The task is motivated by a large body of literature on the impact of diacritization coverage on the reading process of Arabic text, and the scarcity of research into systems to automate it.

\paragraph{Context-Contrastive Prediction}
By considering contextual information alongside non-contextual predictions, our approach builds on the role of diacritics in text, particularly in the context of the Arabic script as used by humans where accuracy is not the only consideration.
The result is a simple, efficient, and configurable method which can be integrated with any existing diacritization system. It opens doors to enhanced diacritization accuracy and selection. The indicators we introduce provide valuable signals for evaluating such advancements.

\paragraph{The Effect of Diacritics on Reading}
A significant motivation for our work rests upon prior research that into the substantial influence of diacritization on the reading process of Arabic text.
Multiple studies have consistently shown that skilled readers tend to read extensively diacritized text at a slower pace than undiacritized text. This phenomenon, supported by research involving diverse age groups and linguistic backgrounds, has significant implications for reading accuracy and comprehension \citep{Haitham16, Ibrahim2013ReadingIA, aula14, Midhwah2020ArabicDA, ROMAN1987431, Hermena15, VariationDiacritics22}.
Other studies also suggest that diacritization not only affects reading speed but also directly impacts reading accuracy \citep{ROMAN1987431, Hermena15}. In this work, we complement this large body of research by conducting a behavioral experiment that utilized our $\fDiacPartial$ approach to partially diacritize news headlines from the \dataset{WikiNews} test set. Our results show that partially diacritized text is easier to read than fully diacritized text, and in some cases can lead to better understanding than no diacritization at all.

\paragraph{Implications}
An exciting potential for this work is optimizing the reading experience for different groups of readers with different needed levels of diacritization, including people with dyslexia and visual impairments.
Since diacritization plays an important role in disambiguating the meaning of the text, it is crucial to intelligently select the ones that aid readability and comprehension of Arabic text without excessive marking; making it accessible and accommodating to a broader audience. This aligns with the broader goals of inclusivity and accessibility in NLP applications.

\paragraph{Future Directions} 
Further research is needed to gather high-quality benchmark data for partial diacritization, to enable more traditional and direct performance metrics alongside the indicators we propose. One promising direction is evaluating the method on other languages which utilize diacritical marks in their orthographies.

\section{Conclusion}
In conclusion, our $\fDiacPartial$ approach for partial diacritization using context-contrastive prediction contributes to the field of diacritization in an area with far-reaching implications for enhancing the reading experience of Arabic text. Grounded in previous research on the influence of diacritization on reading and further supported by behavioral experiments conducted in this work---our approach paves the way for advancements that benefit readers of diverse backgrounds and abilities. Since our approach integrates seamlessly with existing Arabic diacritization systems, it can be used by models trained on domain-specific text as well. We also propose a battery of performance indicators to gauge the competence of partial diacritization systems using fully-diacritized test data to mitigate the lack of publicly available benchmarks. Finally, we introduce \modelname{TD2}---a Transformer adaptation of the \modelname{D2} model which offers a different performance profile as shown by our proposed indicators. As Arabic NLP continues to evolve, our approach serves as a promising direction for enhancing diacritization and its impact on text accessibility and comprehension.

\section*{Limitations}
The method presented in this work is tested on a single human language. While we believe it should be able to generalize to other languages that need similar diacritics restoration, its utility needs further observation. In particular, it may be the case that partial diacritization patterns in languages besides Arabic are different in such a way that the premise of this work no longer holds (wherein ease of guessing is regarded as a major factor in omitting a diacritical mark).

In addition, this method is deliberately kept simple, which builds on an over-simplified view of the human task. Readers guess diacritics when reading, but they may do so while incorporating context via some simple or fast mechanisms. Whether that is at a deep level equivalent to guessing the reading of a single word in isolation remains to be seen. Our method uses non-contextual application, $\fWord$, as a proxy for this process, which is likely relatively close in performance to the same mechanism in a human. Nevertheless, we make no claim that this is indeed the natural mechanism, nor that the proxy exhibits identical performance distribution to the natural mechanism.



\bibliographystyle{acl_natbib}
\bibliography{anthology}

\appendix

\section*{Appendix}

\section{Experimenting with [Transformer] D2}


\subsection{Datasets}
In this work, we train on data from the \dataset{Tashkeela} \citep{fadel19} and \dataset{Ashaar} \citep{ashaar} corpora. We test on the \dataset{Tashkeela} testset.
The \dataset{Tashkeela} corpus has been collected mostly from Islamic classical books \cite{tashkeela17} and contains mostly classical Arabic sentences.
Ashaar is a corpus of Arabic poetry verses covering poems from different eras.
\BTableref{tab:table-data-sizes} shows the number of tokens for the \dataset{Tashkeela} and Ashaar\footnote{The reported numbers reflect dataset cleaning to keep only Arabic letters and diacritics.} datasets.

\begin{table}[h]
\centering
\begin{tabular}{lrrr}
\toprule
                   & \textbf{Train} & \textbf{Dev} & \textbf{Test} \\ 
\midrule
\textbf{\dataset{Tashkeela}} & $2,462,695$    & $120,190$    & $125,343$ \\ 
\textbf{\dataset{Ashaar}}    & $793,181$      & $46,055$     & $39,023$ \\
\bottomrule
\end{tabular}
\caption{Token Counts for Each Data Split in the \dataset{Tashkeela} and Ashaar Diacritization Datasets}
\label{tab:table-data-sizes}
\end{table}

\subsection{Majority Voting using a Sliding Window}
\label{sec:majority-voting}
Following prior work that utilizes an overlapping context window approach with a voting mechanism to enhance diacritic prediction for individual characters \citep{mubarak19-highly}, we segment each input sentence into multiple overlapping windows. We present each segment to the model separately. This approach has proven effective, as localized context often contains enough information for accurate predictions. We use similar segmentation parameters to \citet{alkhamissi-etal-2020-deep}
\paragraph{Training}
For the training and validation sets we use a window of $10$ words and a stride of $2$.
\paragraph{Inference}
The same character may appear in different contexts, and therefore potentially result in different diacritized forms. We implement a popularity voting mechanism to narrow down the prediction. When a tie arises, we randomly select one of the top options. Testing is done with a sliding window of $20$ words at a stride of $2$.

\subsection{Datapath of D2, TD2}
The token model's output is averaged per word to result in exactly one feature vector (instead of one for each sub-word).
This aggregated vector $z^w$ is concatenated with the character embedding of each character in the word and down-projected to the model feature size, to get character input $x^{c,w}$.
The character encoder transforms the inputs $\{ x^{c,w} \mid c \in \func{chars}(w) \}$ into character feature vectors $z^{c,w}$ which are passed to the final classifier.

\paragraph{Training}
The full model is tuned using the AdamW optimizer \citep{adamw} with $0.2$ dropout and $\SI{5e-4}{}$ LR, and follows a linear schedule of $500$ warmup steps and $10,000$ total training steps. The best checkpoint at $1,000$ step intervals is picked by the combined \dataset{Tashkeela} and \dataset{Ashaar} dev sets.

\section{Demo and Examples}
\label{sec:demo}
\begin{figure}[ht]
\centering
\begin{subfigure}[t]{0.98\linewidth}
    \includegraphics[width=1.0\linewidth]{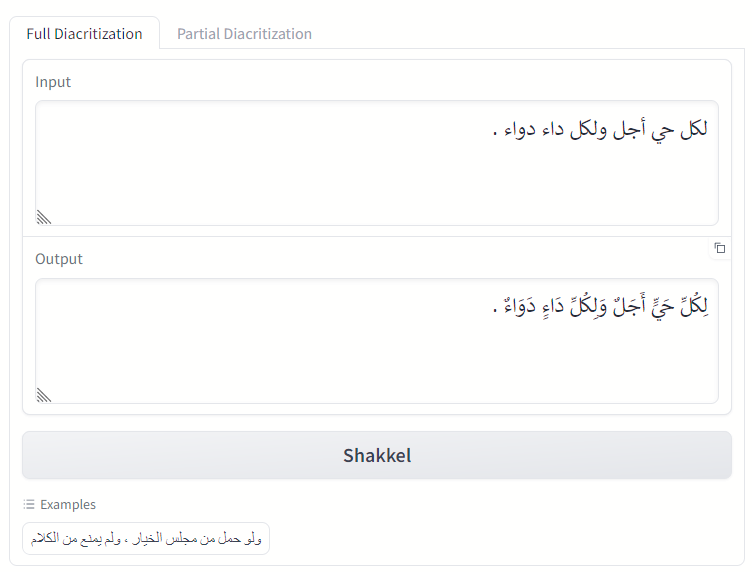}
    \caption{Full Diacritization UI: All predicted diacritics are returned. The model is capable of explicitly predicting no diacritics on a letter, but we did not notice this often.}
    \label{fig:hf-demo-full-diac}
\end{subfigure}\\
\begin{subfigure}[t]{0.98\linewidth}
    \includegraphics[width=1.0\linewidth]{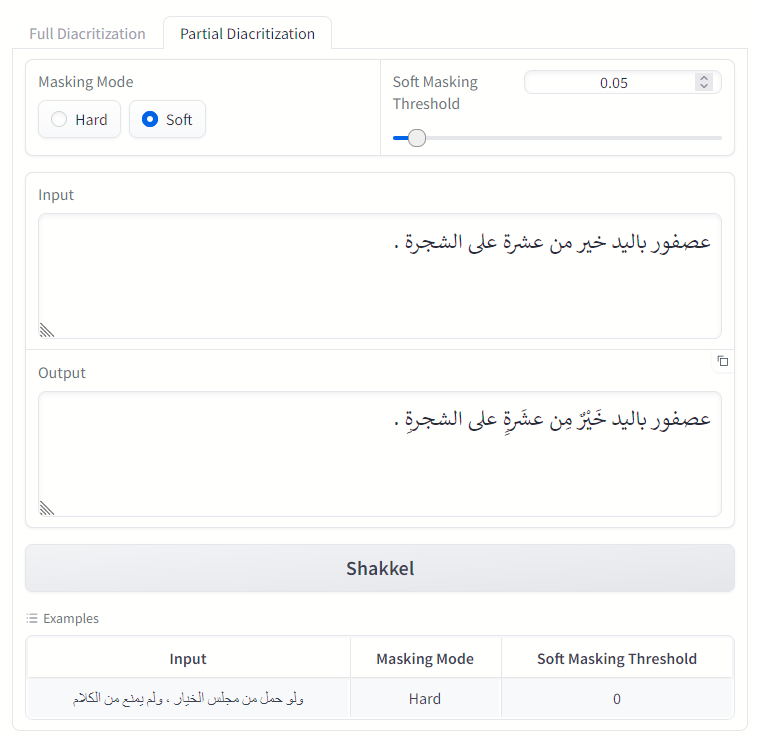}
    \caption{Partial Diacritization UI: Hard mode and Soft mode with threshold. This allows some rough control over the coverage percentage of output.}
    \label{fig:hf-demo-partial-diac}
\end{subfigure}
\caption{Demo UI hosted by HuggingFace supporting Full (Top) and Partial (Bottom) diacritization modes.}
\label{fig:hf-demo}
\end{figure}

We developed an online Demo on Huggingface\footnotemark{} which allows choice between Full Diacritization, and Partial Diacritization with Hard and Soft modes. See \cref{fig:hf-demo-full-diac,fig:hf-demo-partial-diac}. The demo supports the \modelname{D2}/\modelname{TD2} models (trained on \dataset{Tashkeela} and \dataset{Tashkeela}/\dataset{Ashaar}). Since they focus mostly on classical Arabic text, users are advised not to anticipate optimal performance when applying this model to Modern Standard Arabic (MSA) or informal Arabic dialects.
\footnotetext{\href{https://huggingface.co/spaces/bkhmsi/Partial-Arabic-Diacritization}{huggingface.co/spaces/bkhmsi/Partial-Arabic-Diacritization}}

\subsection{More Partial Diacritization Examples}
The following examples are lines taken from a poem by Rami Mohamed (2016) called ``We Will Stay Here''. We applied the models as indicated in the System column. The input text included no diacritics at all. See \cref{tab:expanded_examples}.


\begin{table*}[ht]
\centering
\begin{tabular}{llcr}
\toprule
    \# &
    System \hspace{2em} &
    Text/Output &
    Error Count
\\ \midrule
    & Truth
        & \aratext{سَوْفَ نَبْقَى هُنَا .. كَيْ يَزُولَ الْأَلَمُ}
    & \\
    & Translation
        & We will Stay Here .. That the Pain may one day Cease
    & \\
1 & \modelname{TD2}
    (Full)
    & \aratext{سَوْفَ نَبْقَى هُنَا .. كَيْ يَزُولَ الأَلَمُ}
    & 0 \\
1 & \quad
    (MV, Hard)
    & \aratext{سوفَ نبقى هُنا .. كيْ يزولَ الألمُ}
    & 0 \\
3 & \modelname{D2}
    (Full)
    & \aratext{سَوْفَ نَبْقَى هُنَا .. كَيْ يَزُولَ الْأَلَمُ}
    & 0 \\
4 &
    \quad (MV, hard)
    & \aratext{سوف نبقى هنا .. كي يزولَ الألمُ}
    & 0 \\
5 &
    \quad (SP, soft > 0.05)
    & \aratext{سَوف نبقى هنا .. كي يزولَ الألمُ}
    & 0 \\
6 &
    \quad (SP, soft > 0.01)
    & \aratext{سَوف نبقَى هنا .. كيْ يزولَ الألمُ}
    & 0 \\
\midrule
    & Truth
        & \aratext{فَلْنَقُمْ كُلُّنَا .. بِالدَّوَاءِ وَالْقَلَمِ}
    & \\
    & Translation
        & Let Us all Rise .. With Healing and Writing
    & \\
7 & \modelname{TD2}
    (Full)
    & \aratext{فَلْنَقُمْ كُلُّنَا .. بِالدَّوَاءِ وَالْقَلَمِ}
    & 0 \\
8 & \quad
    (MV, Hard)
    & \aratext{فلْنَقُمْ كلُّنا .. بالدواء والقلمِ}
    & 0 \\
9 & \modelname{D2}
    (Full)
    & \aratext{فَلْنَقِمْ كُلّنَا .. بِالدَّوَاءِ وَالْقَلَمِ}
    & 1 \\
10&
    \quad (MV, hard)
    & \aratext{فلنقم كلّنا .. بالدواء والقلمِ}
    & 0 \\
11&
    \quad (SP, soft > 0.01)
    & \aratext{فلْنقم كلّنا .. بالدواء والقلم}
    & 0 \\
\midrule
    & Truth
        & \aratext{فَرْحَتِي وَصَرْخَتِي .. تَكَادُ تُسْمِعُ الْأَصَمَّ}
    & \\
    & Translation
        & My Joys and Screams .. Through Deafness Nearly Heard
    & \\
12& \modelname{TD2}
    (Full)
    & \aratext{فَرْحَتِي وَصَرْخَتِي .. تَكَادُ تَسْمَعُ الأَصَمَّ}
    & 2 \\
13& \quad
    (MV, Hard)
    & \aratext{فرْحتي وصرختي .. تكادُ تسمع الأصمَّ}
    & 0 \\
14& \modelname{D2}
    (Full)
    & \aratext{فَرْحَتِي وَصُرْخَتِي .. تَكَادُ تَسْمَعُ الْأَصَمَّ}
    & 3 \\
15&
    \quad (MV, hard)
    & \aratext{فرحتي وصرختِي .. تكاد تَسمع الأصمَّ}
    & 0 \\
16 &
    \quad (SP, soft > 0.01)
    & \aratext{فَرحتِي وصرختِي .. تكَادُ تَسمعُ الأصم}
    & 1 \\
\bottomrule
\end{tabular}
\caption{
Examples use \modelname{D2} from \citet{alkhamissi-etal-2020-deep}.
Notably, examples (6, 8, 11, 15) result in outputs similar to a native speaker's, aside from the rare error.
For examples \# 11 and 16, soft thresholds higher or lower than 0.01 did not change the output appreciably. The examples used are lines from the poem ``We Will Stay Here'' by Rami Mohamed (2016) (``\aratext{سوف نبقى هنا}'' by \aratext{رامي محمد}). All examples are generated via the demo (\cref{sec:demo}).
}


\label{tab:expanded_examples}
\end{table*}






\end{document}